\pdfoutput=1

\documentclass[11pt]{article}

\usepackage[final]{acl}

\usepackage{times}
\usepackage{latexsym}

\usepackage[most]{tcolorbox}

\usepackage[T1]{fontenc}

\usepackage[utf8]{inputenc}

\usepackage{microtype}

\usepackage{inconsolata}

\usepackage{adjustbox}
\usepackage{graphicx}
\usepackage{dsfont}
\usepackage{amsmath}
\usepackage{booktabs}
\usepackage{array}
\usepackage{multirow}
\usepackage[ruled,vlined]{algorithm2e}

\usepackage{hyperref}

\usepackage{pifont}
\newcommand{\cmark}{\ding{51}} 
\newcommand{\xmark}{\ding{55}} 
\usepackage{booktabs} 
\usepackage{makecell}

\usepackage{float}     
\title{Semantic Token Clustering for Efficient Uncertainty Quantification in Large Language Models}

\author{Qi Cao, Andrew Gambardella, Takeshi Kojima, Yutaka Matsuo, Yusuke Iwasawa\\
         The University of Tokyo, Japan\\
         \texttt{\url{qi.cao@weblab.t.u-tokyo.ac.jp}}}

\begin{document}
\maketitle
\begin{abstract}
Large language models (LLMs) have demonstrated remarkable capabilities across diverse tasks. However, the truthfulness of their outputs is not guaranteed, and their tendency toward overconfidence further limits reliability. Uncertainty quantification offers a promising way to identify potentially unreliable outputs, but most existing methods rely on repeated sampling or auxiliary models, introducing substantial computational overhead. To address these limitations, we propose Semantic Token Clustering (STC), an efficient uncertainty quantification method that leverages the semantic information inherently encoded in LLMs. Specifically, we group tokens into semantically consistent clusters using embedding clustering and prefix matching, and quantify uncertainty based on the probability mass aggregated over the corresponding semantic cluster. Our approach requires only a single generation and does not depend on auxiliary models. Experimental results show that STC achieves performance comparable to state-of-the-art baselines while substantially reducing computational overhead.\footnote{Code will be available at \url{https://github.com/ccqq77/semantic_token_clustering}.}
\end{abstract}

\section{Introduction}
Large language models (LLMs) achieve impressive performance across diverse tasks but still fail to guarantee factual accuracy, which is a critical limitation, especially in high-stakes domains such as healthcare, law, and science. Their tendency to generate plausible-sounding yet incorrect responses further complicates error detection, underscoring the need for effective uncertainty quantification to identify and manage unreliable outputs.

A natural approach is to allow LLMs to explicitly express their uncertainty verbally. However, due to the well-known overconfidence problem~\cite{xiong2024can}, LLMs often exhibit high confidence even when their responses are plausible but incorrect. Recent studies have attempted to address this issue by quantifying uncertainty in natural language generation, for example, by sampling multiple generations, leveraging external natural language inference (NLI) models to estimate the semantic relationships among them, and measuring uncertainty using semantic dispersion~\cite{kuhn2023semantic,farquhar2024detecting,lin2024generating}.

Despite their effectiveness, most prior approaches require repeated sampling or auxiliary models~\cite{kuhn2023semantic,farquhar2024detecting,lin2024generating}, introducing substantial computational overhead and failing to fully exploit the semantic structure encoded in the LLM's internal representations. In this work, we propose \textit{Semantic Token Clustering (STC)}, a novel and efficient approach for uncertainty quantification that directly leverages internal semantic representations, thereby eliminating the need for external models and multiple generations. Our method achieves performance comparable to state-of-the-art baselines while substantially reducing computational overhead.
Our method offers three key advantages:

\noindent\textbf{Leveraging internal representations.} The method employs token embedding clustering to link the internal semantic representations of LLMs with uncertainty quantification, enabling more effective use of their inherent semantic structure.

\noindent\textbf{Easy and self-contained implementation.} The method requires no fine-tuning, supervised data collection, or external models, relying solely on unsupervised uncertainty quantification in a self-contained manner. It can therefore be readily applied to any off-the-shelf white-box LLM.

\noindent\textbf{Computational efficiency.} Our approach quantifies uncertainty from a single generation, and computationally intensive steps such as embedding clustering are performed offline, minimizing inference-time overhead. It is particularly suitable for resource-constrained and low-latency scenarios.

\section{Related Work}

Existing uncertainty quantification methods for LLMs can be broadly categorized into supervised and unsupervised approaches. Supervised methods typically train additional probes to predict the correctness of generations~\citep{azaria-mitchell-2023-internal, liu2024uncertainty}. However, these methods require labeled data and additional training, and they are not guaranteed to generalize to out-of-distribution data, which limits their flexibility and applicability.

In contrast, unsupervised methods quantify uncertainty directly from model outputs, logits, or internal states without additional training. Logit-based metrics, such as Perplexity~\citep{fomicheva-etal-2020-unsupervised}, compute uncertainty scores directly from token-level logits. 
Sampling-based methods such as Semantic Entropy~\citep{kuhn2023semantic,farquhar2024detecting}, EigenScore~\citep{chen2023inside}, and various semantic dispersion metrics~\citep{lin2024generating} quantify uncertainty by measuring the semantic diversity/consistency across multiple stochastic generations.
Closely related to our work, Claim Conditioned Probability (CCP)~\citep{fadeeva-etal-2024-fact} quantifies token-level uncertainty from a single generation but relies on an NLI model, incurring significant computational overhead.

Despite their effectiveness, existing unsupervised methods either overlook semantic consistency or rely on multiple generations and external models. In contrast, our approach directly leverages semantic information inherently encoded in LLMs, enabling efficient and self-contained uncertainty quantification from a single generation. Table~\ref{table:motiv} summarizes the key differences between our method and existing baselines.

\begin{table}[hb] 
  \centering
  \small
  \caption{Key differences between the proposed uncertainty quantification method and existing methods.}
\label{table:motiv}
  \begin{adjustbox}{max width=1.0\columnwidth}
  \begin{tabular}{@{}lccccc@{}} 
    \toprule
      & \thead{Semantic\\Aware}
      & \thead{Single\\Sample}
      & \thead{External\\Model\\Free}
      & \thead{Overhead}\\
      \midrule
    \hyperlink{cite.fomicheva-etal-2020-unsupervised}{Perplexity}           & \xmark & \cmark & \cmark & Low  \\ 
    \hyperlink{cite.kadavath2022language}{P(True)}              & \xmark & \cmark & \cmark & Medium  \\ 
    \hyperlink{cite.lindley1956measure}{Predictive Entropy}   & \xmark  & \xmark & \cmark & High \\ 
    \hyperlink{cite.malinin2021uncertainty}{LN Entropy}           & \xmark  & \xmark & \cmark & High \\ 
    \hyperlink{cite.duan-etal-2024-shifting}{TokenSAR} & \cmark & \cmark & \xmark & Medium \\ 
    \hyperlink{cite.wang-etal-2024-conu}{ConU}     & \cmark & \xmark & \xmark & High \\ 
    \hyperlink{cite.kuhn2023semantic}{Semantic Entropy}     & \cmark & \xmark & \xmark & High \\ 
    \hyperlink{cite.lin2024generating}{Ecc}     & \cmark & \xmark & \xmark & High \\ 
    \hyperlink{cite.lin2024generating}{EigV}    & \cmark & \xmark & \xmark & High \\ 
    \hyperlink{cite.lin2024generating}{Deg}     & \cmark & \xmark & \xmark & High \\ 
    \hyperlink{cite.chen2023inside}{EigenScore} & \cmark & \xmark & \cmark & High \\ 
    \hyperlink{cite.duan-etal-2024-shifting}{SentenceSAR} & \cmark & \xmark & \xmark & High \\ 
    \hyperlink{cite.duan-etal-2024-shifting}{SAR}     & \cmark & \xmark & \xmark & High \\ 
    \hyperlink{cite.fadeeva-etal-2024-fact}{CCP}     & \cmark & \cmark & \xmark & High \\ 
    Ours  & \cmark & \cmark & \cmark & Low \\  
\bottomrule
  \end{tabular}
  \end{adjustbox}
\end{table}

\section{Problem}
In this study, we focus on uncertainty quantification in LLMs for specific generations. Specifically, given an input prompt $x$ and a generated response $y$, the goal is to estimate a score aligned with the risk that the response $y$ is incorrect. Formally, the uncertainty estimate can be expressed as:
\begin{equation}
\mathcal{U}(x, y) = g\!\left(\hat{p}\big(C=0 \mid x, y\big)\right),
\label{eq:expected}
\end{equation}
where $C$ is a binary correctness indicator, and $g$ is a monotonically increasing link function.

In this study, we aim to develop a computationally efficient method to quantify uncertainty directly from a single generation.

\begin{figure}[t]
  \centering
    \includegraphics[width=0.49\textwidth]{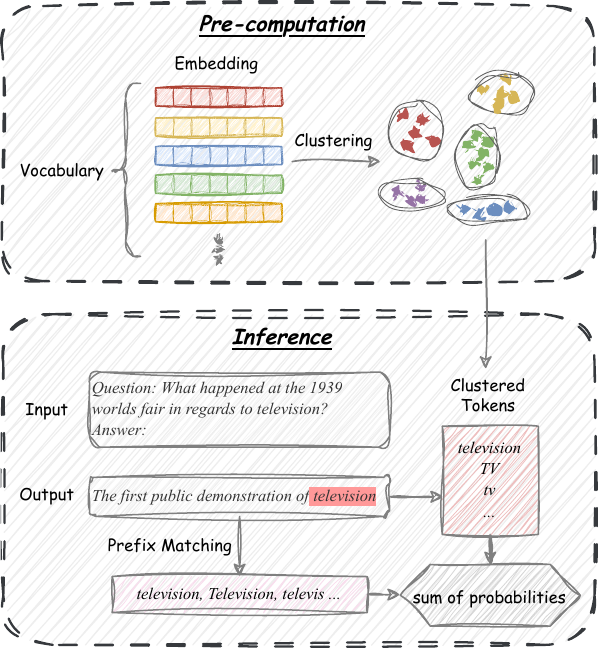}
  \caption{Illustration of the proposed method. Token embedding clustering is performed in the pre-computation stage. During inference, we aggregate next-token probability mass over embedding-clustered and prefix-matched tokens to quantify uncertainty.
  }
  \label{fig:framework}
  
\end{figure}

\begin{figure*}[ht]
  \centering
    \includegraphics[width=0.9\textwidth]{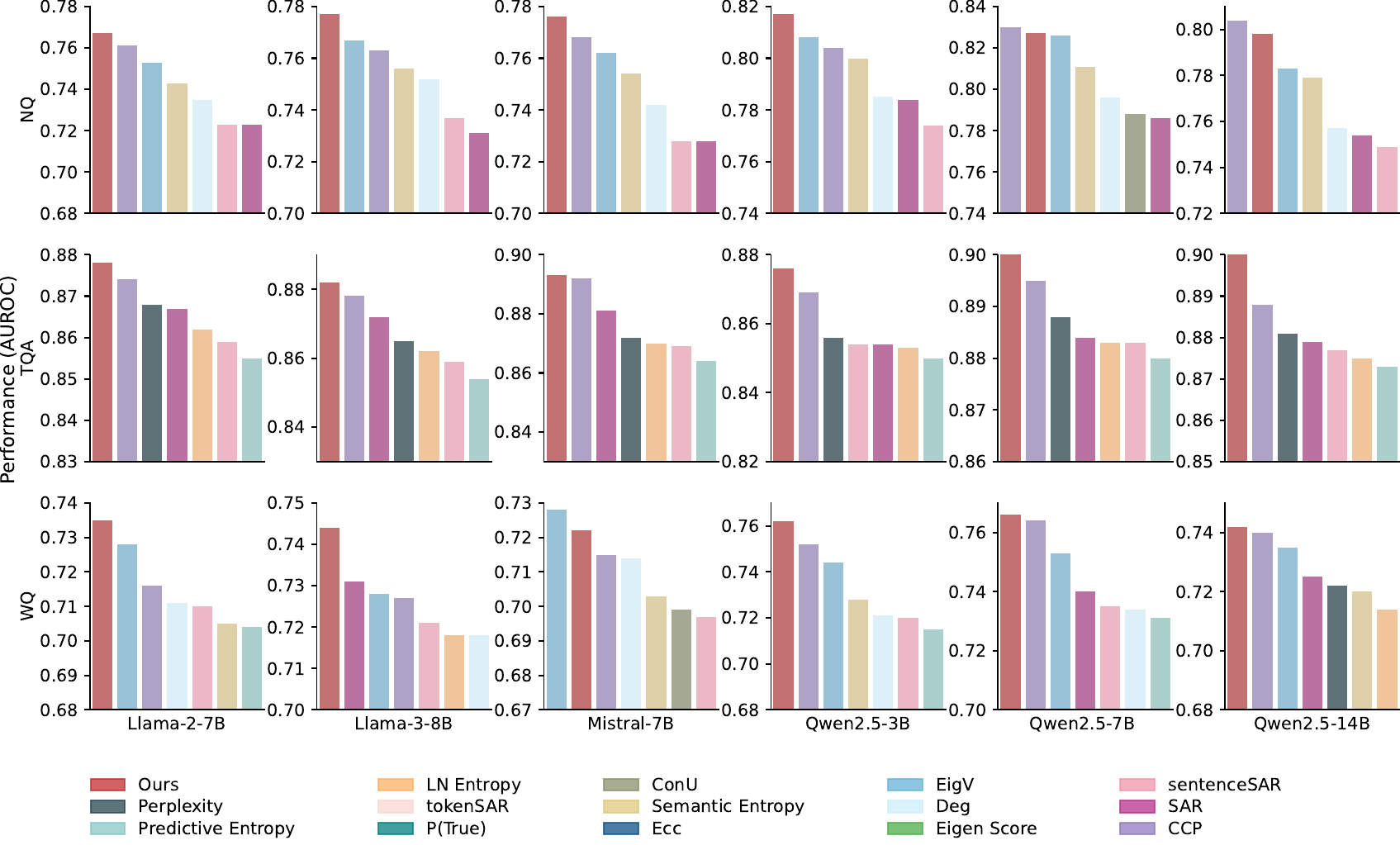}
  \caption{Performance comparison between our method and baseline approaches across different models and datasets. For clarity, only the top seven methods ranked by performance are shown in each subfigure, with methods ordered by descending performance. Metric: AUROC.}
  \label{fig:bar}
  
\end{figure*}

\begin{figure*}[ht]
  \centering
    \includegraphics[width=0.85\textwidth]{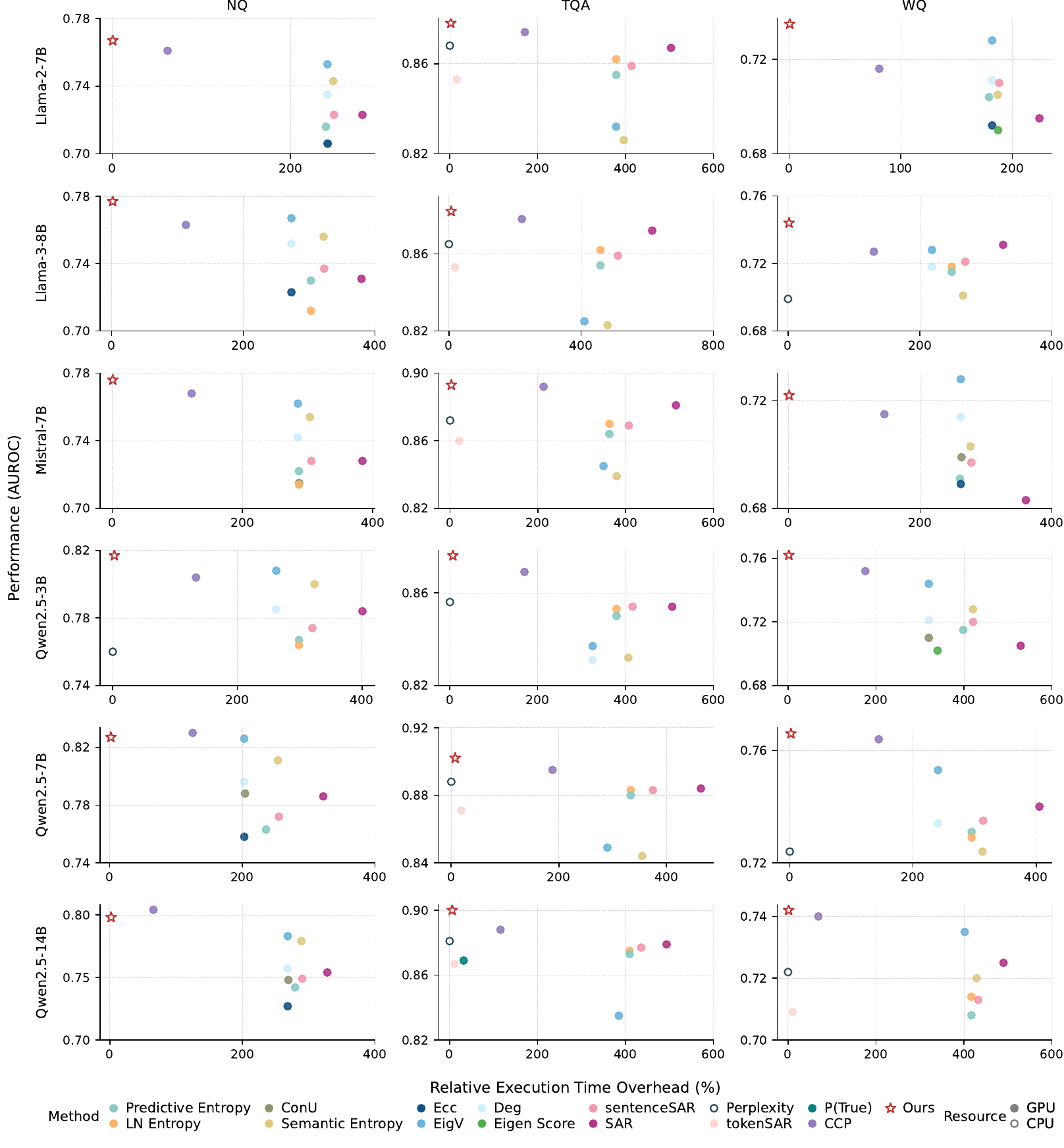}
  \caption{Efficiency comparison across methods. The performance (AUROC) and relative execution time overhead (\%) are plotted on the y-axis and x-axis, respectively, illustrating the efficiency of the proposed method. The relative execution time overhead represents the additional execution time required for uncertainty quantification relative to basic inference. For clarity, only the top ten methods ranked by performance are shown in each subfigure.}
  \label{fig:efficiency}
  
\end{figure*}

\section{Methodology}

Our method quantifies uncertainty using the model's next-token probability distribution at each decoding step, as it directly reflects uncertainty in token selection. Since probability mass is often distributed across multiple semantically consistent tokens (e.g., ``TV'' vs. ``television''), the probability of a single token may underestimate the model's confidence. To address this, we cluster candidate tokens based on semantic similarity and aggregate their probability mass within each cluster to obtain a cluster-based estimate.

As shown in Figure~\ref{fig:framework}, our method measures token-level uncertainty by grouping tokens using embedding clustering and prefix matching. The uncertainty score is computed through a two-stage process:

\paragraph{Pre-computation Stage.}
Inspired by recent work on text embedding, in particular LENS~\cite{lei-etal-2025-enhancing}, we group tokens into semantically consistent clusters based on their embeddings using an unsupervised clustering algorithm, such as Agglomerative Clustering~\cite{mullner2011modern} (implementation details are provided in Appendix~\ref{app:implementations}). Examples of these clusters are shown in Appendix~\ref{app:examples}. The clustering is performed offline, enabling the resulting clusters to be directly used during inference without introducing additional computational overhead.

\paragraph{Inference Stage.}
During inference, we aggregate token probabilities within each semantic cluster at every decoding step to compute an uncertainty score. Because tokenization does not always align with meaningful semantic units, individual tokens may lack sufficient semantic information when considered in isolation. To address this, we incorporate additional semantic information from subsequent context through prefix matching. Specifically, we check whether a candidate token serves as a prefix of the subsequent generation. For example, regardless of whether the subsequent generation is ``television'' as a single token or split into ``tele'' and ``vision'', the tokens ``television'', ``tele'', and ``televis'' are all considered prefix-matched. This process enhances the semantic consistency of clusters by grouping tokens that remain consistent with the subsequent generation.

Formally, the embedding-clustered token set is defined as
\begin{equation}
\mathcal{T}_i^{\mathit{e}} = \{\, t \in \mathcal{V} \mid \text{cluster}(t) = \text{cluster}(y_i) \,\},
\label{eq:cluster_set}
\end{equation}
where $\mathcal{T}_i^{\mathit{e}}$ denotes the set of tokens identified through embedding clustering, $\mathcal{V}$ denotes the model vocabulary, and $\text{cluster}(\cdot)$ maps each token to its corresponding semantic cluster.

Similarly, the prefix-matched token set is defined as
\begin{equation}
\mathcal{T}_i^{\mathit{p}} = \{\, t \in \mathcal{V} \mid \text{norm}(y_{i:})~\mathrel{\text{startswith}}~\text{norm}(t) \,\},
\label{eq:prefix_set}
\end{equation}
where $\mathcal{T}_i^{\mathit{p}}$ denotes the set of tokens identified through prefix matching, $\text{norm}(\cdot)$ performs case- and space-insensitive normalization, and $y_{i:}$ denotes the substring of the remaining output sequence starting from position $i$.

At each decoding step, the clustered probability mass is computed by aggregating the probabilities of all semantically consistent tokens:
\begin{equation}
\hat{p}_c(y_i \mid x, y_{<i}) = \sum_{t \in \mathcal{T}_i} p(t \mid x, y_{<i}),
\label{eq:clustered_likelihood}
\end{equation}
where $\mathcal{T}_i = \mathcal{T}_i^{\mathit{e}} \cup \mathcal{T}_i^{\mathit{p}}$ denotes the union of tokens identified through embedding clustering and prefix matching at step $i$.

The overall uncertainty score for the generated sequence is estimated as one minus the product of the clustered probability masses across decoding steps:
\begin{equation}
\mathcal{S}(x,y) = 1 - \prod_{i=1}^{n} \hat{p}_c(y_i \mid x,y_{<i}),
\label{eq:uncertainty_score}
\end{equation}

Overall, our approach provides an efficient and self-contained pipeline for uncertainty quantification, leveraging the semantic information inherently encoded in LLMs without relying on external models or multiple sampled generations.

\begin{table*}[t]
\centering
\caption{Ablation study results. Metric: AUROC.}
\begin{adjustbox}{max width=1.8\columnwidth}
\begin{tabular}{llcccccc}
\toprule
Dataset & Method & Llama-2-7B & Llama-3-8B & Mistral-7B & Qwen2.5-3B & Qwen2.5-7B & Qwen2.5-14B\\
\midrule
\multirow[]{4}{*}{NQ}  
& Ours & \textbf{0.767} & \textbf{0.777} & \textbf{0.776} & \textbf{0.817} & \textbf{0.827} & \textbf{0.798} \\
& w/o embedding clustering & 0.744 & 0.746 & 0.751 & 0.788 & 0.777 & 0.754 \\
& w/o prefix matching & 0.761 & 0.776 & 0.773 & \textbf{0.817} & 0.826 & 0.795 \\
& probability & 0.731 & 0.743 & 0.739 & 0.785 & 0.770 & 0.748 \\
\cline{1-8}
\multirow[]{4}{*}{TQA} 
& Ours & \textbf{0.878} & \textbf{0.882} & \textbf{0.893} & \textbf{0.876} & \textbf{0.902} & \textbf{0.900} \\
& w/o embedding clustering & 0.876 & 0.874 & 0.884 & 0.873 & 0.899 & 0.894 \\
& w/o prefix matching & 0.874 & 0.876 & 0.887 & 0.873 & 0.899 & 0.896 \\
& probability & 0.867 & 0.861 & 0.871 & 0.864 & 0.891 & 0.882 \\

\cline{1-8}
\multirow[]{4}{*}{WQ}  
 & Ours & \textbf{0.735} & \textbf{0.744} & \textbf{0.722} & \textbf{0.762} & \textbf{0.766} & \textbf{0.742} \\
& w/o embedding clustering & 0.718 & 0.733 & 0.708 & 0.744 & 0.756 & 0.728 \\
& w/o prefix matching & 0.722 & 0.736 & 0.713 & 0.757 & 0.764 & 0.734 \\
& probability & 0.702 & 0.716 & 0.695 & 0.731 & 0.747 & 0.718 \\

\bottomrule
\end{tabular}
\end{adjustbox}
\label{table:ablation}
\end{table*}

\section{Experiments}

\subsection{Setup}

\paragraph{Baselines.} We compare our proposed method with baselines, including single-generation methods such as Perplexity~\cite{fomicheva-etal-2020-unsupervised}, tokenSAR~\cite{duan-etal-2024-shifting}, and CCP~\cite{fadeeva-etal-2024-fact}; sampling-based methods such as Predictive Entropy~\cite{lindley1956measure}, LN-Entropy~\cite{malinin2021uncertainty}, EigenScore~\cite{chen2023inside}, ConU~\cite{wang-etal-2024-conu}, Semantic Entropy~\cite{kuhn2023semantic}, Ecc, EigV, and Deg~\cite{lin2024generating}, as well as sentenceSAR and SAR~\cite{duan-etal-2024-shifting}; and the prompting-based method P(True)~\cite{kadavath2022language}.

\paragraph{Models.} We conduct experiments using open-source models: Llama-2-7B~\cite{touvron2023llama}, Llama-3-8B~\cite{llama3modelcard}, Mistral-7B-v0.3~\cite{jiang2023mistral} and Qwen2.5 models with 3B, 7B, and 14B parameters~\cite{qwen2025qwen25technicalreport}.\footnote{We access and utilize the weights and configurations of these models via HuggingFace: \url{https://huggingface.co/}.}

\paragraph{Datasets.} We evaluate the methods on three datasets: TriviaQA (TQA;~\citealp{joshi-etal-2017-triviaqa}), Natural Questions (NQ;~\citealp{kwiatkowski-etal-2019-natural}), and WebQuestions (WQ;~\citealp{berant-etal-2013-semantic}). For TQA and NQ, we follow the preprocessing settings in~\citet{lin2024generating}, while for WQ, we use the original test set. The resulting numbers of processed samples are 9,960 for TQA, 3,610 for NQ, and 2,032 for WQ.

\paragraph{Details.} Further details regarding evaluation and implementation specifics are provided in Appendix~\ref{app:evaluation} and Appendix~\ref{app:implementations}, respectively.

\subsection{Results}

Figure~\ref{fig:bar} compares the performance of different methods. Our approach achieves performance comparable to state-of-the-art baselines across datasets and models, demonstrating its effectiveness. Figure~\ref{fig:efficiency} visualizes the trade-off between performance and overhead to illustrate efficiency. Our method is positioned in the upper-left corner, indicating that beyond strong empirical results, it is also highly efficient, with substantially lower computational overhead than state-of-the-art baselines. In particular, compared with CCP, our approach achieves competitive performance while reducing inference-time overhead by an average of 98\%.

Our approach requires neither multiple generations nor external models, and embedding clustering is performed offline during pre-computation (the overhead of this stage is discussed in Appendix~\ref{sec:pre_compute_overhead}). Consequently, the inference-time overhead is minimized, and the process can be executed efficiently on CPUs without requiring GPUs.

\subsection{Ablation Study}

We conduct ablation studies to evaluate the contributions of key components in our method, with the results presented in Table~\ref{table:ablation}. Removing either embedding clustering or prefix matching individually leads to moderate performance degradation. This occurs because the two components are complementary: embedding clustering captures semantic similarity, whereas prefix matching captures surface-form consistency. In many cases, the embedding-clustered tokens already include those identified by prefix matching (e.g., ``television'' and ``Television'' belong to the same embedding cluster and are also prefix-matched). When both components are removed, the method reduces to computing the probability of the generated response, resulting in a substantial performance drop. These findings underscore the effectiveness of both embedding clustering and prefix matching in our uncertainty quantification method. A detailed sensitivity analysis on the embedding clustering is provided in Appendix~\ref{sec:sensitivity_clustering}.

\section{Conclusion}
We propose \textit{Semantic Token Clustering (STC)}, an efficient method for uncertainty quantification in LLMs that leverages their inherent semantic information. Our approach quantifies uncertainty from a single generation without requiring multiple generations or external models. Experimental results show that STC achieves performance comparable to state-of-the-art baselines while substantially reducing computational overhead, demonstrating both its effectiveness and efficiency.

\section{Limitations}
First, the proposed method requires access to token logits and token embeddings, which are typically unavailable in closed-source models. Consequently, users cannot directly apply our method to such models without access to these internal representations.

Second, the proposed method currently relies on static token embeddings and semantic relationships derived from the LLM's vocabulary. Although these embeddings inherently encode rich semantic information, they may introduce noise into clusters due to their context-independent nature. A potential source of this noise is polysemy, which may cause tokens with divergent meanings to be grouped into the cluster when contextual information is not considered. In practice, LLMs tend to assign very low probabilities to tokens with incompatible meanings, which may help mitigate this issue to some extent. Nevertheless, incorporating more context-aware semantic representations (e.g., contextualized embeddings) could reduce such noise and further enhance the performance and robustness of the method. Future work may explore integrating these context-aware representations to improve the reliability and informativeness of uncertainty quantification.

Finally, similar to CCP~\cite{fadeeva-etal-2024-fact}, the proposed method does not explicitly address the calibration of uncertainty scores. Nevertheless, it could be post-calibrated if needed.

\bibliography{anthology-1,anthology-2,custom}

\appendix
\renewcommand{\thesection}{\Alph{section}}

\section{Examples of Clusters}
\label{app:examples}
Table~\ref{table:example} demonstrates example clusters obtained from Llama-3-8B~\cite{llama3modelcard}.

\begin{table}[ht]
  \centering
  \caption{Example clusters from Llama-3-8B. The symbol ``Ġ'' represents a space character in its tokenizer.}
  \label{table:example}
  \begin{adjustbox}{max width=0.999\columnwidth}
  \begin{tabular}{@{}lc@{}} 
    \toprule
      \thead{Cluster Examples}\\
      \midrule
      TV, tv, ĠTV, Ġtv, ĠTelevision, Ġtelevision, Ġtelevis ...\\
      \midrule
      Beautiful, ĠBeautiful, ĠGorgeous, Ġgorgeous ...\\
      \midrule
      Plane, planes, Ġplane, Ġairplane, ĠAircraft, Ġaircraft ...\\
      \midrule
      Possible, ĠPossible, Ġconceivable, Ġimaginable ...\\
      \midrule
      Market, ĠMarkets, ĠMarketplace, Ġmarketplace, \_market ...\\
      \midrule
      Cold, cold, ĠCold, Ġcold, Ġchilly, Ġchilling, Ġchilled ...\\
      \midrule
      Buy, ĠBuy, ĠBought, ĠPurchase, Ġpurchase, Ġpurchased ...\\
      \midrule
      Trash, trash, ĠTrash, Ġtrash, Ġjunk, Ġgarbage ...\\

\bottomrule
  \end{tabular}
  \end{adjustbox}
  
\end{table}

\section{Evaluation}
\label{app:evaluation}

We evaluate the effectiveness of our proposed method in quantifying uncertainty for deterministic responses generated via greedy decoding. This setting closely reflects real-world scenarios, particularly when querying factual information from large language models (LLMs). For comparison, we also implement sampling-based baselines by generating auxiliary responses using temperature sampling to estimate uncertainty.

To assess uncertainty quantification methods, we first determine the correctness of each generated answer, which serves as ground truth for evaluating uncertainty estimation quality. Previous studies have commonly used metrics such as ROUGE-L~\cite{lin-2004-rouge} and semantic similarity~\cite{reimers-gurevych-2019-sentence} to measure answer correctness by comparing generated responses to reference answers. However, these metrics are often unreliable: reference answers may not cover all valid responses, and heuristic thresholds are not universally applicable. Moreover, recent work~\cite{santilli2024on} has shown spurious interactions between uncertainty scores and these evaluation metrics, further undermining their reliability.

To address these limitations, we employ GPT-4.1 (2025-04-14 version)~\cite{openai2025gpt41} as an evaluator to determine answer correctness. We prompt GPT-4.1 to assess factual accuracy directly, rather than strict adherence to reference answers. Using these correctness labels, we evaluate uncertainty quantification performance via the area under the receiver operating characteristic curve (AUROC). This approach provides a robust and reliable assessment, directly measuring the ability of uncertainty scores to distinguish between correct and incorrect responses without relying on heuristic thresholds.

\section{Implementation}
\label{app:implementations}
To compute our proposed uncertainty score, we first cluster token embeddings into semantically consistent groups in the pre-computation stage. Specifically, we concatenate each token's input embeddings (from token embedding layer) and output embeddings (from language modeling head) to form unified semantic representations, which are then clustered using Agglomerative Clustering~\cite{mullner2011modern} with cosine distance as distance measure. We use the scikit-learn implementation~\cite{scikit-learn}. Based on the empirical findings in LENS~\cite{lei-etal-2025-enhancing}, we set the number of clusters to 16,000. Following CCP~\cite{fadeeva-etal-2024-fact}, we exclude function words using the NLTK stopword list~\cite{bird-loper-2004-nltk,bird2009natural}. In addition, tokens representing Arabic numerals are omitted from embedding clustering, since numerals with similar embeddings are not necessarily mathematically equivalent. This pre-computation step significantly reduces computational overhead and eliminates the need for GPU resources during uncertainty quantification at inference time.

To ensure a fair comparison with sampling-based approaches, we generate five additional responses per question using temperature sampling (temperature = 0.5), resulting in six generations per question (one deterministic response via greedy decoding and five sampled responses). Flash Attention 2~\cite{dao2023flashattention2} is employed during sampling to improve efficiency. The temperature value of 0.5 is chosen following prior work, as it was found to be optimal for baseline methods such as Semantic Entropy~\cite{kuhn2023semantic} and EigenScore~\cite{chen2023inside}. While our method does not require multiple generations or utilize information from these additional samples, we generate them solely to compute uncertainty scores for sampling-based baselines. No inference-time intervention techniques are applied, and all experiments use the original model weights and activations.

For computational resources, methods requiring GPU acceleration are run on a node with two Intel Xeon Platinum 8368 CPUs and eight Nvidia A100 GPUs (40GB each). Methods that do not require GPU acceleration, including ours, are executed on a node with the same CPUs but without GPUs.

\section{Detailed Results}

\label{sec:add_results}
Table~\ref{app:table:AUROC}, Table~\ref{app:table:PRR}, Table~\ref{app:table:absolute_time}, and Table~\ref{app:table:relarive_time} present detailed results for AUROC performance, Prediction Rejection Ratio (PRR) performance (following LM-Polygraph Benchmark~\cite{vashurin-etal-2025-benchmarking}), absolute execution time at inference, and relative execution time overhead at inference, respectively.

\section{Pre-computation Overhead}
\label{sec:pre_compute_overhead}

Embedding clustering in the pre-computation stage is performed using the scikit-learn implementation~\cite{scikit-learn}, with pairwise distances computed in PyTorch~\cite{NEURIPS2019_bdbca288}. The execution time depends on the vocabulary size and embedding dimension. As shown in Table~\ref{table:pre_overhead}, for Qwen2.5-14B~\cite{qwen2025qwen25technicalreport}, the model with the largest vocabulary size (152,064) and embedding dimension (5,120) in our experiments, the execution time remains acceptable, as the algorithm needs to be executed only once per model.

\begin{table}[ht]
\centering
\caption{Execution time of embedding clustering in the pre-computation stage for different models.}
\label{table:pre_overhead}
\begin{tabular}{lc}
\toprule
Model & Time (mm:ss) \\
\midrule
Llama-2-7B & 01:08 \\
Llama-3-8B & 18:24 \\
Mistral-7B & 00:48 \\
Qwen2.5-3B & 23:46 \\
Qwen2.5-7B & 28:05 \\
Qwen2.5-14B & 33:46 \\
\bottomrule
\end{tabular}
\end{table}

\begin{table*}[t]
\centering
\caption{Full experimental results for performance comparison across methods. Metric: AUROC.}
\label{app:table:AUROC}
\resizebox{\textwidth}{!}{
\begin{tabular}{llcccccc}
\toprule
Dataset & Method & Llama-2-7B & Llama-3-8B & Mistral-7B&  Qwen2.5-3B & Qwen2.5-7B & Qwen2.5-14B\\
\midrule
\multirow[]{5}{*}{NQ}  
& Perplexity & 0.697 & 0.705 & 0.712 & 0.760 & 0.744 & 0.708 \\
& Predictive Entropy & 0.716 & 0.730 & 0.722 & 0.767 & 0.763 & 0.742 \\
& LN Entropy & 0.702 & 0.712 & 0.714 & 0.764 & 0.755 & 0.721 \\
& tokenSAR & 0.673 & 0.682 & 0.683 & 0.736 & 0.724 & 0.708 \\
& P(True) & 0.550 & 0.546 & 0.488 & 0.680 & 0.727 & 0.725 \\
& ConU & 0.706 & 0.708 & 0.715 & 0.759 & 0.788 & 0.748 \\
& Semantic Entropy & 0.743 & 0.756 & 0.754 & 0.800 & 0.811 & 0.779 \\
& Ecc & 0.706 & 0.723 & 0.714 & 0.753 & 0.758 & 0.727 \\
& EigV & 0.753 & \underline{0.767} & 0.762 & \underline{0.808} & 0.826 & 0.783 \\
& Deg & 0.735 & 0.752 & 0.742 & 0.785 & 0.796 & 0.757 \\
& Eigen Score & 0.691 & 0.691 & 0.684 & 0.750 & 0.747 & 0.716 \\
& sentenceSAR & 0.723 & 0.737 & 0.728 & 0.774 & 0.772 & 0.749 \\
& SAR & 0.723 & 0.731 & 0.728 & 0.784 & 0.786 & 0.754 \\
& CCP & \underline{0.761} & 0.763 & \underline{0.768} & 0.804 & \textbf{0.830} & \textbf{0.804} \\
& Ours & \textbf{0.767} & \textbf{0.777} & \textbf{0.776} & \textbf{0.817} & \underline{0.827} & \underline{0.798} \\

\hline
\multirow[]{15}{*}{TQA} 
& Perplexity & 0.868 & 0.865 & 0.872 & 0.856 & 0.888 & 0.881 \\
& Predictive Entropy & 0.855 & 0.854 & 0.864 & 0.850 & 0.880 & 0.873 \\
& LN Entropy & 0.862 & 0.862 & 0.870 & 0.853 & 0.883 & 0.875 \\
& tokenSAR & 0.853 & 0.853 & 0.860 & 0.831 & 0.871 & 0.867 \\
& P(True) & 0.533 & 0.640 & 0.587 & 0.739 & 0.832 & 0.869 \\
& ConU & 0.750 & 0.753 & 0.758 & 0.749 & 0.771 & 0.754 \\
& Semantic Entropy & 0.826 & 0.823 & 0.839 & 0.832 & 0.844 & 0.832 \\
& Ecc & 0.799 & 0.792 & 0.810 & 0.809 & 0.815 & 0.802 \\
& EigV & 0.832 & 0.825 & 0.845 & 0.837 & 0.849 & 0.835 \\
& Deg & 0.824 & 0.818 & 0.837 & 0.831 & 0.839 & 0.827 \\
& Eigen Score & 0.807 & 0.805 & 0.809 & 0.820 & 0.832 & 0.815 \\
& sentenceSAR & 0.859 & 0.859 & 0.869 & 0.854 & 0.883 & 0.877 \\
& SAR & 0.867 & 0.872 & 0.881 & 0.854 & 0.884 & 0.879 \\
& CCP & \underline{0.874} & \underline{0.878} & \underline{0.892} & \underline{0.869} & \underline{0.895} & \underline{0.888} \\
& Ours & \textbf{0.878} & \textbf{0.882} & \textbf{0.893} & \textbf{0.876} & \textbf{0.902} & \textbf{0.900} \\
   
\hline
\multirow[]{15}{*}{WQ} 
& Perplexity & 0.664 & 0.699 & 0.651 & 0.673 & 0.724 & 0.722 \\
& Predictive Entropy & 0.704 & 0.715 & 0.691 & 0.715 & 0.731 & 0.708 \\
& LN Entropy & 0.684 & 0.718 & 0.665 & 0.694 & 0.729 & 0.714 \\
& tokenSAR & 0.634 & 0.684 & 0.614 & 0.643 & 0.694 & 0.709 \\
& P(True) & 0.529 & 0.560 & 0.562 & 0.625 & 0.715 & 0.707 \\
& ConU & 0.681 & 0.678 & 0.699 & 0.710 & 0.722 & 0.703 \\
& Semantic Entropy & 0.705 & 0.701 & 0.703 & 0.728 & 0.724 & 0.720 \\
& Ecc & 0.692 & 0.680 & 0.689 & 0.691 & 0.699 & 0.672 \\
& EigV & \underline{0.728} & 0.728 & \textbf{0.728} & 0.744 & 0.753 & 0.735 \\
& Deg & 0.711 & 0.718 & 0.714 & 0.721 & 0.734 & 0.708 \\
& Eigen Score & 0.690 & 0.693 & 0.682 & 0.702 & 0.709 & 0.684 \\
& sentenceSAR & 0.710 & 0.721 & 0.697 & 0.720 & 0.735 & 0.713 \\
& SAR & 0.695 & \underline{0.731} & 0.683 & 0.705 & 0.740 & 0.725 \\
& CCP & 0.716 & 0.727 & 0.715 & \underline{0.752} & \underline{0.764} & \underline{0.740} \\
& Ours & \textbf{0.735} & \textbf{0.744} & \underline{0.722} & \textbf{0.762} & \textbf{0.766} & \textbf{0.742} \\

\bottomrule
\end{tabular}}
\end{table*}

\begin{table*}[ht]
\centering
\caption{Full experimental results for performance comparison across methods. Metric: PRR.}
\label{app:table:PRR}
\resizebox{\textwidth}{!}{
\begin{tabular}{llcccccc}
\toprule
Dataset & Method & Llama-2-7B & Llama-3-8B & Mistral-7B&  Qwen2.5-3B & Qwen2.5-7B & Qwen2.5-14B\\
\midrule
\multirow[]{15}{*}{NQ}  
& Perplexity & 0.442 & 0.496 & 0.508 & 0.502 & 0.513 & 0.460 \\
& Predictive Entropy & 0.500 & 0.563 & 0.537 & 0.559 & 0.573 & 0.564 \\
& LN Entropy & 0.462 & 0.519 & 0.514 & 0.536 & 0.545 & 0.501 \\
& tokenSAR & 0.415 & 0.446 & 0.455 & 0.483 & 0.490 & 0.491 \\
& P(True) & 0.111 & 0.153 & -0.058 & 0.383 & 0.512 & 0.495 \\
& ConU & 0.381 & 0.376 & 0.406 & 0.413 & 0.499 & 0.451 \\
& Semantic Entropy & 0.460 & 0.480 & 0.497 & 0.541 & 0.577 & 0.530 \\
& Ecc & 0.428 & 0.473 & 0.455 & 0.488 & 0.512 & 0.477 \\
& EigV & 0.481 & 0.545 & 0.521 & 0.572 & 0.600 & 0.534 \\
& Deg & 0.461 & 0.516 & 0.502 & 0.539 & 0.567 & 0.495 \\
& Eigen Score & 0.406 & 0.439 & 0.395 & 0.464 & 0.497 & 0.421 \\
& sentenceSAR & 0.510 & \underline{0.573} & \underline{0.546} & 0.571 & 0.586 & 0.573 \\
& SAR & 0.506 & 0.548 & 0.531 & \underline{0.574} & 0.596 & 0.568 \\
& CCP & \underline{0.513} & 0.552 & \underline{0.546} & 0.565 & \underline{0.647} & \underline{0.643} \\
& Ours & \textbf{0.541} & \textbf{0.599} & \textbf{0.591} & \textbf{0.613} & \textbf{0.650} & \textbf{0.645} \\
\hline
\multirow[]{15}{*}{TQA} 
& Perplexity & 0.805 & 0.811 & 0.822 & 0.768 & 0.834 & 0.827 \\
& Predictive Entropy & 0.789 & 0.797 & 0.812 & 0.767 & 0.821 & 0.817 \\
& LN Entropy & 0.797 & 0.807 & 0.820 & 0.767 & 0.826 & 0.819 \\
& tokenSAR & 0.785 & 0.795 & 0.803 & 0.726 & 0.805 & 0.806 \\
& P(True) & 0.070 & 0.387 & 0.203 & 0.521 & 0.739 & 0.810 \\
& ConU & 0.510 & 0.498 & 0.504 & 0.432 & 0.512 & 0.491 \\
& Semantic Entropy & 0.653 & 0.651 & 0.673 & 0.644 & 0.669 & 0.671 \\
& Ecc & 0.622 & 0.635 & 0.664 & 0.625 & 0.645 & 0.639 \\
& EigV & 0.668 & 0.683 & 0.691 & 0.660 & 0.693 & 0.677 \\
& Deg & 0.653 & 0.659 & 0.681 & 0.653 & 0.692 & 0.675 \\
& Eigen Score & 0.584 & 0.614 & 0.579 & 0.646 & 0.671 & 0.637 \\
& sentenceSAR & 0.794 & 0.804 & 0.819 & 0.772 & 0.825 & 0.822 \\
& SAR & \underline{0.806} & \underline{0.820} & 0.832 & 0.764 & 0.823 & 0.825 \\
& CCP & \underline{0.806} & 0.814 & \underline{0.842} & \underline{0.779} & \underline{0.839} & \underline{0.828} \\
& Ours & \textbf{0.823} & \textbf{0.836} & \textbf{0.854} & \textbf{0.799} & \textbf{0.855} & \textbf{0.861} \\    
\hline
\multirow[]{15}{*}{WQ} 
& Perplexity & 0.399 & 0.452 & 0.388 & 0.331 & 0.519 & 0.539 \\
& Predictive Entropy & 0.496 & 0.533 & 0.500 & 0.495 & 0.560 & 0.539 \\
& LN Entropy & 0.463 & 0.530 & 0.443 & 0.442 & 0.543 & 0.532 \\
& tokenSAR & 0.328 & 0.415 & 0.299 & 0.310 & 0.475 & 0.515 \\
& P(True) & 0.090 & 0.170 & 0.168 & 0.280 & 0.503 & 0.473 \\
& ConU & 0.362 & 0.292 & 0.428 & 0.401 & 0.434 & 0.424 \\
& Semantic Entropy & 0.399 & 0.381 & 0.418 & 0.439 & 0.455 & 0.433 \\
& Ecc & 0.416 & 0.396 & 0.448 & 0.393 & 0.419 & 0.431 \\
& EigV & 0.500 & 0.462 & \underline{0.523} & 0.470 & 0.501 & 0.520 \\
& Deg & 0.462 & 0.455 & 0.490 & 0.425 & 0.491 & 0.475 \\
& Eigen Score & 0.414 & 0.450 & 0.445 & 0.369 & 0.434 & 0.426 \\
& sentenceSAR & 0.507 & \underline{0.543} & 0.512 & 0.504 & 0.568 & 0.548 \\
& SAR & 0.463 & 0.540 & 0.466 & 0.469 & 0.562 & 0.546 \\
& CCP & \underline{0.512} & 0.512 & 0.505 & \underline{0.528} & \underline{0.602} & \underline{0.553} \\
& Ours & \textbf{0.568} & \textbf{0.570} & \textbf{0.540} & \textbf{0.573} & \textbf{0.628} & \textbf{0.587} \\
\bottomrule
\end{tabular}}
\end{table*}

\begin{table*}[ht]
\centering
\caption{Full experimental results on absolute execution time overhead at inference. Metric: seconds.}
\label{app:table:absolute_time}
\resizebox{\textwidth}{!}{
\begin{tabular}{llcccccc}
\toprule
Dataset & Method & Llama-2-7B & Llama-3-8B & Mistral-7B&  Qwen2.5-3B & Qwen2.5-7B & Qwen2.5-14B\\
\midrule
\multirow[]{15}{*}{NQ}  
& Perplexity & \textbf{0.129} & \textbf{0.064} & \textbf{0.075} & \textbf{0.108} & \textbf{0.173} & \textbf{0.148} \\
& Predictive Entropy & 1373.943 & 1006.703 & 1028.696 & 824.862 & 863.792 & 1717.797 \\
& LN Entropy & 1374.020 & 1006.781 & 1028.779 & 824.943 & 863.880 & 1717.879 \\
& tokenSAR & 50.865 & 49.245 & 68.746 & 51.428 & 71.280 & 59.225 \\
& P(True) & 41.669 & 41.258 & 50.507 & 25.471 & 47.472 & 61.176 \\
& ConU & 1387.134 & 910.635 & 1030.129 & 713.243 & 747.901 & 1655.025 \\
& Semantic Entropy & 1420.822 & 1070.526 & 1089.431 & 894.150 & 929.426 & 1775.456 \\
& Ecc & 1383.779 & 908.374 & 1023.511 & 724.153 & 743.618 & 1648.152 \\
& EigV & 1383.420 & 908.361 & 1023.446 & 724.462 & 743.579 & 1648.646 \\
& Deg & 1383.066 & 907.995 & 1023.109 & 723.760 & 743.213 & 1647.769 \\
& Eigen Score & 1407.099 & 951.202 & 1077.065 & 731.121 & 746.425 & 1735.322 \\
& sentenceSAR & 1425.529 & 1073.833 & 1097.110 & 884.191 & 934.741 & 1783.411 \\
& SAR & 1608.375 & 1262.202 & 1378.736 & 1106.083 & 1179.924 & 2015.596 \\
& CCP & 355.393 & 375.405 & 437.571 & 368.500 & 458.807 & 403.897 \\
& Ours & \underline{3.028} & \underline{5.767} & \underline{3.250} & \underline{7.081} & \underline{7.235} & \underline{12.127} \\
\hline
\multirow[]{15}{*}{TQA}
& Perplexity & \textbf{0.175} & \textbf{0.243} & \textbf{0.215} & \textbf{0.299} & \textbf{0.226} & \textbf{0.173} \\
& Predictive Entropy & 1001.054 & 858.933 & 766.178 & 1185.428 & 799.185 & 1414.291 \\
& LN Entropy & 1001.284 & 859.173 & 766.410 & 1185.655 & 799.403 & 1414.523 \\
& tokenSAR & 42.683 & 33.505 & 44.933 & 57.409 & 45.261 & 39.718 \\
& P(True) & 77.504 & 64.794 & 84.098 & 42.733 & 67.394 & 110.757 \\
& ConU & 1043.482 & 823.381 & 792.341 & 1042.695 & 738.870 & 1381.948 \\
& Semantic Entropy & 1046.829 & 899.641 & 801.556 & 1269.717 & 850.439 & 1449.723 \\
& Ecc & 1000.627 & 768.468 & 738.256 & 1016.296 & 696.236 & 1329.495 \\
& EigV & 1000.469 & 768.285 & 738.037 & 1015.869 & 695.372 & 1329.348 \\
& Deg & 999.541 & 767.505 & 737.141 & 1014.938 & 694.200 & 1328.461 \\
& Eigen Score & 1039.766 & 800.555 & 779.427 & 1063.181 & 719.904 & 1426.132 \\
& sentenceSAR & 1093.151 & 958.088 & 859.242 & 1299.645 & 897.532 & 1505.557 \\
& SAR & 1330.471 & 1152.862 & 1086.359 & 1581.633 & 1111.684 & 1705.063 \\
& CCP & 451.485 & 413.634 & 449.906 & 530.589 & 451.137 & 400.007 \\
& Ours & \underline{6.156} & \underline{12.810} & \underline{6.138} & \underline{21.703} & \underline{17.417} & \underline{20.265} \\
\hline
\multirow[]{15}{*}{WQ} 
& Perplexity & \textbf{0.099} & \textbf{0.039} & \textbf{0.058} & \textbf{0.090} & \textbf{0.109} & \textbf{0.128} \\
& Predictive Entropy & 1177.434 & 693.584 & 880.289 & 1226.562 & 1033.338 & 1303.567 \\
& LN Entropy & 1177.483 & 693.630 & 880.338 & 1226.606 & 1033.386 & 1303.611 \\
& tokenSAR & 79.666 & 42.500 & 83.959 & 77.041 & 79.275 & 33.112 \\
& P(True) & 45.502 & 44.397 & 47.685 & 24.845 & 43.000 & 34.523 \\
& ConU & 1203.438 & 618.553 & 888.340 & 985.693 & 844.796 & 1266.870 \\
& Semantic Entropy & 1228.517 & 741.462 & 934.841 & 1295.211 & 1095.089 & 1340.643 \\
& Ecc & 1195.505 & 609.604 & 884.749 & 985.680 & 842.305 & 1256.113 \\
& EigV & 1195.448 & 609.569 & 884.717 & 985.672 & 842.272 & 1256.075 \\
& Deg & 1195.290 & 609.385 & 884.519 & 985.459 & 842.112 & 1255.907 \\
& Eigen Score & 1230.437 & 628.671 & 937.786 & 1046.944 & 885.305 & 1350.655 \\
& sentenceSAR & 1237.058 & 751.045 & 939.009 & 1295.818 & 1098.173 & 1352.041 \\
& SAR & 1473.864 & 911.192 & 1218.985 & 1628.814 & 1417.933 & 1531.791 \\
& CCP & 530.564 & 363.803 & 492.417 & 541.178 & 505.802 & 215.823 \\
& Ours & \underline{3.106} & \underline{4.408} & \underline{3.265} & \underline{5.967} & \underline{6.650} & \underline{4.228} \\
\bottomrule
\end{tabular}}
\end{table*}

\begin{table*}[ht]
\centering
\caption{Full experimental results on relative execution time overhead at inference. Metric: \%.}
\label{app:table:relarive_time}
\resizebox{\textwidth}{!}{
\begin{tabular}{llcccccc}
\toprule
Dataset & Method & Llama-2-7B & Llama-3-8B & Mistral-7B&  Qwen2.5-3B & Qwen2.5-7B & Qwen2.5-14B\\

\midrule

\multirow[]{15}{*}{NQ}  
& Perplexity & \textbf{0.023} & \textbf{0.019} & \textbf{0.021} & \textbf{0.039} & \textbf{0.047} & \textbf{0.024} \\
& Predictive Entropy & 239.879 & 302.765 & 286.607 & 298.737 & 235.837 & 279.658 \\
& LN Entropy & 239.893 & 302.789 & 286.630 & 298.766 & 235.861 & 279.672 \\
& tokenSAR & 8.881 & 14.811 & 19.153 & 18.625 & 19.461 & 9.642 \\
& P(True) & 7.275 & 12.408 & 14.072 & 9.225 & 12.961 & 9.960 \\
& ConU & 242.183 & 273.873 & 287.006 & 258.312 & 204.196 & 269.439 \\
& Semantic Entropy & 248.064 & 321.960 & 303.529 & 323.831 & 253.757 & 289.045 \\
& Ecc & 241.597 & 273.193 & 285.163 & 262.264 & 203.027 & 268.320 \\
& EigV & 241.534 & 273.189 & 285.145 & 262.375 & 203.016 & 268.400 \\
& Deg & 241.472 & 273.079 & 285.050 & 262.121 & 202.916 & 268.258 \\
& Eigen Score & 245.668 & 286.073 & 300.083 & 264.787 & 203.793 & 282.511 \\
& sentenceSAR & 248.886 & 322.955 & 305.668 & 320.224 & 255.208 & 290.340 \\
& SAR & 280.809 & 379.607 & 384.133 & 400.586 & 322.149 & 328.140 \\
& CCP & 62.049 & 112.903 & 121.913 & 133.458 & 125.266 & 65.755 \\
& Ours & \underline{0.529} & \underline{1.734} & \underline{0.905} & \underline{2.564} & \underline{1.975} & \underline{1.974} \\

\hline
\multirow[]{15}{*}{TQA}
& Perplexity & \textbf{0.066} & \textbf{0.130} & \textbf{0.102} & \textbf{0.096} & \textbf{0.094} & \textbf{0.050} \\
& Predictive Entropy & 378.796 & 458.126 & 363.029 & 379.401 & 333.478 & 409.382 \\
& LN Entropy & 378.883 & 458.254 & 363.139 & 379.474 & 333.569 & 409.449 \\
& tokenSAR & 16.151 & 17.870 & 21.290 & 18.374 & 18.886 & 11.497 \\
& P(True) & 29.327 & 34.559 & 39.847 & 13.677 & 28.122 & 32.060 \\
& ConU & 394.851 & 439.164 & 375.426 & 333.719 & 308.310 & 400.020 \\
& Semantic Entropy & 396.117 & 479.838 & 379.792 & 406.378 & 354.865 & 419.638 \\
& Ecc & 378.635 & 409.875 & 349.799 & 325.270 & 290.520 & 384.837 \\
& EigV & 378.575 & 409.777 & 349.695 & 325.133 & 290.159 & 384.794 \\
& Deg & 378.223 & 409.361 & 349.271 & 324.835 & 289.671 & 384.538 \\
& Eigen Score & 393.445 & 426.989 & 369.307 & 340.275 & 300.396 & 412.809 \\
& sentenceSAR & 413.645 & 511.011 & 407.124 & 415.957 & 374.516 & 435.800 \\
& SAR & 503.446 & 614.897 & 514.737 & 506.208 & 463.875 & 493.549 \\
& CCP & 170.841 & 220.618 & 213.174 & 169.817 & 188.247 & 115.787 \\
& Ours & \underline{2.329} & \underline{6.833} & \underline{2.908} & \underline{6.946} & \underline{7.268} & \underline{5.866} \\

\hline
\multirow[]{15}{*}{WQ} 
& Perplexity & \textbf{0.015} & \textbf{0.014} & \textbf{0.017} & \textbf{0.029} & \textbf{0.031} & \textbf{0.041} \\
& Predictive Entropy & 179.218 & 248.185 & 260.382 & 398.411 & 295.320 & 416.995 \\
& LN Entropy & 179.225 & 248.202 & 260.396 & 398.425 & 295.334 & 417.009 \\
& tokenSAR & 12.126 & 15.208 & 24.834 & 25.024 & 22.656 & 10.592 \\
& P(True) & 6.926 & 15.887 & 14.105 & 8.070 & 12.289 & 11.043 \\
& ConU & 183.176 & 221.337 & 262.763 & 320.172 & 241.436 & 405.256 \\
& Semantic Entropy & 186.993 & 265.318 & 276.518 & 420.709 & 312.968 & 428.855 \\
& Ecc & 181.968 & 218.135 & 261.701 & 320.168 & 240.725 & 401.815 \\
& EigV & 181.959 & 218.122 & 261.692 & 320.165 & 240.715 & 401.803 \\
& Deg & 181.935 & 218.056 & 261.633 & 320.096 & 240.669 & 401.749 \\
& Eigen Score & 187.285 & 224.957 & 277.389 & 340.067 & 253.014 & 432.058 \\
& sentenceSAR & 188.293 & 268.746 & 277.751 & 420.906 & 313.850 & 432.501 \\
& SAR & 224.337 & 326.052 & 360.565 & 529.070 & 405.235 & 490.001 \\
& CCP & 80.757 & 130.180 & 145.653 & 175.785 & 144.555 & 69.039 \\
& Ours & \underline{0.473} & \underline{1.577} & \underline{0.966} & \underline{1.938} & \underline{1.900} & \underline{1.352} \\

\bottomrule
\end{tabular}}
\end{table*}

\section{Sensitivity Analysis on Embedding Clustering}
\label{sec:sensitivity_clustering}
The unsupervised clustering method used in our study is Agglomerative Clustering~\cite{mullner2011modern}. In the default setting, we employ the concatenation of input and output embeddings as token representations, cosine distance as the distance measure, and 16,000 as the number of clusters. We conduct sensitivity analyses on clustering algorithms (Table~\ref{tab:sen_algo}), distance measures (Table~\ref{tab:sen_dist}), number of clusters (Table~\ref{tab:sen_num_cluster}), embedding types (Table~\ref{tab:sen_embed}), and linkage settings (Table~\ref{tab:sen_linkage1}). In each analysis, the default configuration is listed in the bottom row of the corresponding table.

\begin{table*}[ht]
\centering
\caption{Sensitivity analysis with different clustering algorithms. Metric: AUROC.}
\label{tab:sen_algo}
\begin{adjustbox}{max width=1.9\columnwidth}
\begin{tabular}{llcccccc}
\toprule
Dataset & Algorithms & Llama-2-7B & Llama-3-8B & Mistral-7B & Qwen2.5-3B & Qwen2.5-7B & Qwen2.5-14B\\
\midrule
\multirow[]{2}{*}{NQ}  
& Kmeans & 0.766 & 0.773 & 0.775 & 0.817 & 0.828 & 0.796 \\
& Agglomerative & 0.767 & 0.777 & 0.776 & 0.817 & 0.827 & 0.798 \\

\cline{1-8}

\multirow[]{2}{*}{TQA} 
& Kmeans & 0.878 & 0.882 & 0.892 & 0.876 & 0.901 & 0.899 \\
& Agglomerative & 0.878 & 0.882 & 0.893 & 0.876 & 0.902 & 0.900 \\

\cline{1-8}
\multirow[]{2}{*}{WQ}  
& Kmeans & 0.737 & 0.742 & 0.722 & 0.761 & 0.768 & 0.742 \\
& Agglomerative & 0.735 & 0.744 & 0.722 & 0.762 & 0.766 & 0.742 \\

\bottomrule
\end{tabular}
\end{adjustbox}
\end{table*}

\begin{table*}[ht]
\centering
\caption{Sensitivity analysis with different distance measures. Metric: AUROC.}
\label{tab:sen_dist}
\begin{adjustbox}{max width=1.9\columnwidth}
\begin{tabular}{llcccccc}
\toprule
Dataset & Distance & Llama-2-7B & Llama-3-8B & Mistral-7B & Qwen2.5-3B & Qwen2.5-7B & Qwen2.5-14B\\
\midrule
\multirow[]{2}{*}{NQ}  
& Euclidean & 0.767 & 0.777 & 0.776 & 0.817 & 0.827 & 0.798 \\
& Cosine & 0.767 & 0.777 & 0.776 & 0.817 & 0.827 & 0.798 \\

\cline{1-8}
\multirow[]{2}{*}{TQA} 
& Euclidean & 0.878 & 0.882 & 0.893 & 0.876 & 0.902 & 0.900 \\
& Cosine & 0.878 & 0.882 & 0.893 & 0.876 & 0.902 & 0.900 \\

\cline{1-8}
\multirow[]{2}{*}{WQ}  
& Euclidean & 0.735 & 0.744 & 0.722 & 0.762 & 0.766 & 0.742 \\
& Cosine & 0.735 & 0.744 & 0.722 & 0.762 & 0.766 & 0.742 \\

\bottomrule
\end{tabular}
\end{adjustbox}
\end{table*}

\begin{table*}[ht]
\centering
\caption{Sensitivity analysis with different numbers of clusters. Metric: AUROC.}
\label{tab:sen_num_cluster}
\begin{adjustbox}{max width=1.9\columnwidth}
\begin{tabular}{llcccccc}
\toprule
Dataset & Number & Llama-2-7B & Llama-3-8B & Mistral-7B & Qwen2.5-3B & Qwen2.5-7B & Qwen2.5-14B\\
\midrule
\multirow[]{3}{*}{NQ}  
& 8000 & 0.767 & 0.776 & 0.776 & 0.816 & 0.826 & 0.798 \\
& 12000 & 0.767 & 0.776 & 0.776 & 0.816 & 0.827 & 0.798 \\
& 16000 & 0.767 & 0.777 & 0.776 & 0.817 & 0.827 & 0.798 \\
\cline{1-8}

\multirow[]{3}{*}{TQA} 
& 8000 & 0.878 & 0.882 & 0.893 & 0.875 & 0.902 & 0.900 \\
& 12000 & 0.878 & 0.881 & 0.893 & 0.876 & 0.902 & 0.901 \\
& 16000 & 0.878 & 0.882 & 0.893 & 0.876 & 0.902 & 0.900 \\

\cline{1-8}
\multirow[]{3}{*}{WQ}  
& 8000 & 0.736 & 0.743 & 0.722 & 0.761 & 0.766 & 0.742 \\
& 12000 & 0.736 & 0.743 & 0.722 & 0.762 & 0.766 & 0.743 \\
& 16000 & 0.735 & 0.744 & 0.722 & 0.762 & 0.766 & 0.742 \\

\bottomrule
\end{tabular}
\end{adjustbox}
\end{table*}

\begin{table*}[ht]
\centering
\caption{Sensitivity analysis with different embedding types. Metric: AUROC.}
\label{tab:sen_embed}
\begin{adjustbox}{max width=1.9\columnwidth}
\begin{tabular}{llcccccc}
\toprule
Dataset & Embedding & Llama-2-7B & Llama-3-8B & Mistral-7B & Qwen2.5-3B & Qwen2.5-7B & Qwen2.5-14B\\
\midrule
\multirow[]{3}{*}{NQ}  
& Input & 0.766 & 0.775 & 0.776 & 0.817 & 0.826 & 0.795 \\
& Output & 0.768 & 0.769 & 0.776 & 0.817 & 0.827 & 0.799 \\
& Concatenated & 0.767 & 0.777 & 0.776 & 0.817 & 0.827 & 0.798 \\

\cline{1-8}

\multirow[]{3}{*}{TQA} 
& Input & 0.878 & 0.880 & 0.893 & 0.876 & 0.900 & 0.900 \\
& Output & 0.878 & 0.876 & 0.892 & 0.876 & 0.902 & 0.900 \\
& Concatenated & 0.878 & 0.882 & 0.893 & 0.876 & 0.902 & 0.900 \\

\cline{1-8}
\multirow[]{3}{*}{WQ}  
& Input & 0.736 & 0.744 & 0.723 & 0.762 & 0.768 & 0.741 \\
& Output & 0.736 & 0.743 & 0.722 & 0.762 & 0.769 & 0.742 \\
& Concatenated & 0.735 & 0.744 & 0.722 & 0.762 & 0.766 & 0.742 \\

\bottomrule
\end{tabular}
\end{adjustbox}
\end{table*}

\begin{table*}[ht]
\centering
\caption{Sensitivity analysis with different linkage settings. Metric: AUROC.}
\label{tab:sen_linkage1}
\begin{adjustbox}{max width=1.9\columnwidth}
\begin{tabular}{llcccccc}
\toprule
Dataset & Linkage & Llama-2-7B & Llama-3-8B & Mistral-7B & Qwen2.5-3B & Qwen2.5-7B & Qwen2.5-14B\\
\midrule
\multirow[]{3}{*}{NQ}  
& Single & 0.751 & 0.652 & 0.776 & 0.705 & 0.722 & 0.697 \\
& Average & 0.767 & 0.775 & 0.777 & 0.817 & 0.827 & 0.796 \\
& Complete & 0.767 & 0.777 & 0.776 & 0.817 & 0.827 & 0.798 \\

\cline{1-8}

\multirow[]{3}{*}{TQA} 
& Single & 0.859 & 0.752 & 0.893 & 0.753 & 0.783 & 0.764 \\
& Average & 0.878 & 0.882 & 0.893 & 0.876 & 0.902 & 0.900 \\
& Complete & 0.878 & 0.882 & 0.893 & 0.876 & 0.902 & 0.900 \\

\cline{1-8}
\multirow[]{3}{*}{WQ}  
& Single & 0.721 & 0.645 & 0.724 & 0.719 & 0.696 & 0.657 \\
& Average & 0.736 & 0.744 & 0.722 & 0.761 & 0.765 & 0.742 \\
& Complete & 0.735 & 0.744 & 0.722 & 0.762 & 0.766 & 0.742 \\

\bottomrule
\end{tabular}
\end{adjustbox}
\end{table*}

Except for the ``single'' linkage setting (which uses the minimum distance between all observations of two clusters), all other configurations of the clustering method have minimal impact on the performance of uncertainty quantification. This result demonstrates the insensitivity of our method to clustering hyperparameter settings and the stability of semantic representations in the embedding space, which together enable effective clustering of semantically consistent tokens across different settings.

\newpage
\section{Prompts}
For the generation prompts, we generally follow the settings in~\citet{lin2024generating}. The prompts used for TQA, NQ, and WQ are presented below, respectively.

Additionally, the prompt used for the LLM-as-a-Judge evaluation with GPT-4.1 is provided at the end.

\begin{tcolorbox}[title=Prompt for NQ dataset, colback=white, colframe=black]
\ttfamily\small
Answer these questions:
\\
\\
Question:
\\
who makes up the state council in russia
\\
Answer:
\\
governors and presidents
\\
\\
Question:
\\
when does real time with bill maher come back
\\
Answer:
\\
November 9, 2018
\\
\\
Question:
\\
where did the phrase american dream come from
\\
Answer:
\\
the mystique regarding frontier life
\\
\\
Question:
\\
what do you call a group of eels
\\
Answer:
\\
bed
\\
\\
Question:
\\
who wrote the score for mission impossible fallout
\\
Answer:
\\
Lorne Balfe
\\
\\
Question:
\\
\{Insert Question\}
\\
Answer:
\\
\end{tcolorbox}

\begin{tcolorbox}[title=Prompt for TQA dataset, colback=white, colframe=black]
\ttfamily\small
Answer these questions:
\\
\\
Question:
\\
In Scotland a bothy/bothie is a?
\\
Answer:
\\
House
\\
\\
Question:

\{Insert Question\}

Answer:
\\
\end{tcolorbox}

\begin{tcolorbox}[title=Prompt for WQ dataset, colback=white, colframe=black]
\ttfamily\small
Answer these questions:
\\
\\
Question:
\\
where was the ancient region of mesopotamia?
\\
Answer:
\\
Middle East
\\
\\
Question:

\{Insert Question\}

Answer:
\\
\end{tcolorbox}

\begin{tcolorbox}[title=LLM-as-a-Judge Prompt, colback=white, colframe=black]
\ttfamily\small
\textbf{System Message}:

\# Task

Evaluate whether the proposed answer 
to the question is correct based
on real-world factual knowledge. 
Reference answers are provided to
assist in your evaluation.
\\
\\
\# Output

Respond strictly with a single token:

- `True' if the proposed answer is 
correct.

- `False' if the proposed answer is 
incorrect or only partially correct.
\\
\\
\textbf{User Message}:

Question:

\{Insert Question\}
\\
\\
Reference Answer(s):

\{Insert Reference Answers\}
\\
\\
Proposed Answer:

\{Insert Proposed Answer\}
\\
\\
True/False:
\end{tcolorbox}

\end{document}